\documentclass{article}
\usepackage{spconf,amsmath,graphicx}

\usepackage{algorithm}
\usepackage{algorithmic}
\newcommand{\xvec}{{\bf x}}

\usepackage{booktabs}
\usepackage{amsfonts,amssymb}
\usepackage{multirow}
\usepackage{subfigure}
\usepackage{amsthm}
\usepackage{url}            
\usepackage{amsfonts}       
\usepackage{hyperref}

\title{Vision Transformer-based Adversarial Domain Adaptation}
\name{Yahan Li \qquad Yuan Wu$^*$\thanks{*Corresponding author.}}
\address{School of Artificial Intelligence, Jilin University, Changchun, China}
\begin{document}
%
\maketitle
\begin{abstract}

Unsupervised domain adaptation (UDA) aims to transfer knowledge from a labeled source domain to an unlabeled target domain. The most recent UDA methods always resort to adversarial training to yield state-of-the-art results and a dominant number of existing UDA methods employ convolutional neural networks (CNNs) as feature extractors to learn domain invariant features. Vision transformer (ViT) has attracted tremendous attention since its emergence and has been widely used in various computer vision tasks, such as image classification, object detection, and semantic segmentation, yet its potential in adversarial domain adaptation has never been investigated. In this paper, we fill this gap by employing the ViT as the feature extractor in adversarial domain adaptation. Moreover, we empirically demonstrate that ViT can be a plug-and-play component in adversarial domain adaptation, which means directly replacing the CNN-based feature extractor in existing UDA methods with the ViT-based feature extractor can easily obtain performance improvement. The code is available at \url{https://github.com/LluckyYH/VT-ADA}.

\end{abstract}
\section{Introduction}
\label{sec:intro}

Deep neural networks (DNNs) have delivered remarkable achievements across diverse domains, spanning computer vision (CV), natural language processing (NLP), and speech recognition \cite{goodfellow2016deep}. However, these impressive strides are intrinsically tethered to copious volumes of annotated data \cite{han2022survey}. Collecting such labeled data is perennially laborious and resource-intensive, a challenge that underscores the quest for strategies to extrapolate knowledge effectively from label-rich domains to label-scarce ones \cite{pan2009survey}. Unfortunately, conventional supervised machine learning presupposes training and test data be drawn from identical data distributions, a notion often at odds with real-world scenarios due to domain shifts.

The predicament of domain shift, a formidable challenge, has spawned the discipline of domain adaptation, endeavoring to transmute knowledge from source domains rich in labels to target domains bereft of them. In machine learning, Unsupervised Domain Adaptation (UDA) shoulders the more onerous task of transposing knowledge from labeled source domains to unlabeled target domains \cite{wang2018deep}. Early UDA techniques entail reweighting source instances based on their pertinence to the target domain via human-engineered features \cite{gong2013connecting}. The most recent UDA methodologies traverse a different trajectory by projecting data points from both source and target domains into a shared latent space. They strive to diminish the chasm between the data distributions of source and target domains while nurturing features that are both transferrable and discriminative \cite{wang2018deep}.

The quest to minimize domain discrepancy unfolds along two strategies. One avenue leverages metrics like maximum mean discrepancy (MMD) \cite{yan2017mind} or correlation distances \cite{sun2016deep}. Alternatively, adversarial training, as pioneered by \cite{goodfellow2014generative}, and refined by \cite{ganin2016domain}, deploys a captivating two-player minimax game. Here, a domain discriminator locks horns with a feature extractor, with the former striving to demarcate source features from target ones, and the latter endeavoring to obfuscate the discriminator's task \cite{long2018conditional}.

Most extant UDA approaches lean heavily upon Convolutional Neural Networks (CNNs) \cite{wang2018deep,li2021survey}. However, the recent emergence of the Vision Transformer (ViT) \cite{dosovitskiy2020image} has galvanized the computer vision landscape. In stark contrast to CNNs, which glean insights from local image patches, ViT partitions an image into non-overlapping patches and harnesses self-attention to model long-range dependencies among these patches \cite{dosovitskiy2020image}. This raises a compelling inquiry: Can ViT, which has demonstrated its mettle in diverse computer vision domains \cite{liu2022swin,li2022exploring,strudel2021segmenter}, be harnessed to augment adversarial domain adaptation (ADA)?

Our research embarks upon addressing this very question. We introduce the Vision Transformer-based Adversarial Domain Adaptation (VT-ADA) approach by replacing the CNN-based feature extractor employed in ADA methods with ViT. Notably, our empirical investigations underscore that this substitution is not only facile but also yields tangible enhancements in learning domain-invariant features. We conduct extensive experiments across three UDA benchmarks, wherein the observations unequivocally affirm the efficacy of VT-ADA in augmenting the transferability and discriminability of domain-invariant features. Remarkably, our variant of VT-ADA, which integrates ViT into CDAN \cite{long2018conditional} emerges as a potent contender against state-of-the-art ADA methods.

\section{Related Works}
\label{sec:rel}

\subsection{Unsupervised Domain Adaptation}

Unsupervised domain adaptation (UDA) addresses the domain shift challenge by training a model to transition from a labeled source domain to an unlabeled target domain \cite{wang2018deep,zhou2024survey}. Predominantly, UDA methodologies utilize CNN-based feature extractors and adversarial training to cultivate domain-invariant features that exhibit robust generalization across disparate domains \cite{wu2020dual}. Domain Adversarial Neural Networks (DANNs) integrate a gradient reversal layer and adversarial training within the UDA framework \cite{ganin2016domain}, while Adversarial Discriminative Domain Adaptation (ADDA) employs asymmetric feature extractors to facilitate domain alignment \cite{tzeng2017adversarial}. Additionally, the Cross-Domain Transformer (CDTrans) leverages cross-attention mechanisms to derive common representations shared between the source and target domains \cite{xu2021cdtrans}.

\subsection{Vision Transformer}

The Transformer architecture, initially designed to tackle challenges in NLP, incorporates mechanisms such as word embedding, position embedding, and self-attention \cite{vaswani2017attention}. The adaptation of this architecture for visual tasks led to the development of the ViT, which has revolutionized the field of CV by treating images as sequences of non-overlapping patches \cite{dosovitskiy2020image,chang2023survey}. This approach departs from the traditional CNNs that rely on domain-specific inductive biases, embracing instead the advantages of extensive pre-training and global context modeling. ViT has proven its capability, matching or even surpassing CNNs in tasks like image classification and setting the stage for its broader adoption as a replacement for CNNs across various CV challenges. The growing dominance of ViT-based models in the field is evident \cite{liu2022swin}. In this paper, we explore the potential of ViT as a plug-and-play component within existing ADA frameworks, aiming to enhance the discriminative power and transferability of domain-invariant features.

\subsection{Self-Attention}

The self-attention mechanism is a pivotal component in ViT \cite{dosovitskiy2020image}. Specifically, an image $\mathcal{I}\in\mathbb{R}^{H\times W\times C}$ is transformed into a sequence of flattened 2D patches $\xvec\in\mathbb{R}^{N\times (P^2\cdot C)}$, where $(H, W)$ represents the height and weight dimensions of the original image, $C$ denotes the number of channels, $(P, P)$ specifies the resolution of each image patch, and $N=HW/P^2$ is the number of resulting patches. For the self-attention, these patches are initially projected into three vectors: queries $\mathcal{Q}\in\mathbb{R}^{N\times d_k}$, keys $\mathcal{K}\in\mathbb{R}^{N\times d_k}$, and values $\mathcal{V}\in\mathbb{R}^{N\times d_v}$, with $d_k$ and $d_v$ denoting their respective dimensions. The self-attention output is derived as a weighted sum of the values, where each weight is determined by a compatibility function between a query and its corresponding key. The $N$ patches serve as the inputs for the self-attention module, and the process can be formulated as:

\begin{align}
\label{sa}
    SA(\mathcal{Q},\mathcal{K},\mathcal{V})=softmax(\frac{\mathcal{Q}\mathcal{K}^T}{\sqrt{d_k}})\mathcal{V}
\end{align}

\noindent The self-attention module accentuates the interdependencies among patches within the input image. According to eq.\ref{sa}, multi-head self-attention is employed within the ViT to effectively model long-range dependencies. This configuration is elucidated as follows:

\begin{align}
    MSA(\mathcal{Q},\mathcal{K},\mathcal{V})=Concat(head_1,...,head_k)\mathcal{W}^O
\end{align}

\noindent where $head_i=SA(\mathcal{Q}\mathcal{W}_i^{\mathcal{Q}},\mathcal{K}\mathcal{W}_i^{\mathcal{K}},\mathcal{V}\mathcal{W}_i^{\mathcal{V}})$, $\mathcal{W}_i^{\mathcal{Q}}$, $\mathcal{W}_i^{\mathcal{K}}$, $\mathcal{W}_i^{\mathcal{V}}$ are projections of different heads, $\mathcal{W}^O$ is another mapping function.

\section{The Approach}
\label{sec:Appro}

In this paper, we consider UDA tasks in the following setting: there exist abundant labeled data in the source domain, $D^s=\{(\xvec_i^s,y_i^s)\}_{i=1}^{n_s}$ with $\xvec_i^s\in\mathcal{X}$ and $y_i^s\in\mathcal{Y}$, and a set of unlabeled instances in the target domain, $D^t=\{\xvec_i^t\}_{i=1}^{n_t}$ with $\xvec_i^t\in\mathcal{X}$. The data in the two domains are drawn from different distributions $\mathcal{S}$ and $\mathcal{T}$, but share the same label space. The main objective of UDA is to learn a model $h:\mathcal{X}\to\mathcal{Y}$ that has a good capacity of generalizing on both the source and target domains, indicating that $h$ could yield good classification results on both $D^s$ and $D^t$.

\subsection{Adversarial Domain Adaptation}

The central premise of ADA is to cultivate domain-invariant features that generalize across various domains. Originating with Domain Adversarial Neural Networks (DANNs) \cite{ganin2016domain}, adversarial learning has become prevalent for cultivating feature representations that mitigate domain disparities in UDA. Taking DANN as a prototype, it consists of three main components: a feature extractor $F$, a task-specific classifier $C$, and a binary domain discriminator $D$. The feature extractor $F:\mathcal{X}\to\mathbb{R}^m$ transforms an input instance $\xvec$ from the input space $\mathcal{X}$ into a shared latent space $F(\xvec)\in\mathbb{R}^m$. Historically, ADA methods predominantly employ CNN architecture for the feature extractor \cite{wu2020dual}. The classifier $C:\mathbb{R}^m\to\mathcal{Y}$ translates a feature vector from the shared latent space into the label space $\mathcal{Y}$. The domain discriminator $D:\mathbb{R}^m\to[0,1]$ differentiates source features (domain index 0) from target features (domain index 1) within the latent space. Through adversarial training, $F$ is optimized to confound $D$, enabling DANN to acquire features transferable across domains. Concurrently, $F$ and $C$ are trained to minimize the classification error on labeled source data, thereby enhancing the discriminative power of the features across categories. DANN is formally described as follows:

\begin{align}
    \min_{F,C}\max_{D}\mathcal{L}_c(F,C)+\lambda_d\mathcal{L}_{d}(F,D)
\end{align}

\begin{align}
    \mathcal{L}_c(F,C)=\mathbb{E}_{(\xvec^s,y^s)\sim\mathcal{S}} \ell(C(F(\xvec^s)),y^s)
\end{align}

\begin{equation}
    \begin{aligned}
    \mathcal{L}_{d}(F,D)=&\mathbb{E}_{\xvec^s\sim \mathcal{S}}\log[D(F(\xvec^s))] \\
    +&\mathbb{E}_{\xvec^t\sim\mathcal{T}}\log[1-D(F(\xvec^t))]
    \end{aligned}
\end{equation}

\noindent Where $\ell(\cdot,\cdot)$ is the standard cross-entropy loss, and $\lambda_d$ is a trade-off hyperparameter.

Conditional domain adversarial network (CDAN) exploits the classifier predictions that are believed to convey rich discriminative information to conditional adversarial training \cite{long2018conditional}. Specifically, it conditions the domain discriminator $D$ by using the pseudo-labels provided by the classifier through a multilinear mapping, the conditional adversarial loss can be formulated as:

\begin{equation}
    \begin{aligned}
    \mathcal{L}_{d}(F,D)=&\mathbb{E}_{\xvec^s\sim \mathcal{S}}\log[D[F(\xvec^s)),C(F(\xvec^s)]] \\
    +&\mathbb{E}_{\xvec^t\sim\mathcal{T}}\log[1-D[F(\xvec^t),C(F(\xvec^t))]]
    \end{aligned}
\end{equation}

\noindent where $[\cdot,\cdot]$ represents the concatenation of two vectors.

\subsection{Vision Transformer-based Adversarial Domain Adaptation}

To evaluate the potential of ViT in ADA, we introduce a Vision Transformer-based Adversarial Domain Adaptation (VT-ADA) approach that utilizes the standard ViT architecture \cite{dosovitskiy2020image} as the backbone of the feature extractor. Historically, CNN-based methods have prevailed in UDA research, achieving notable successes as highlighted by \cite{wang2018deep}. In our VT-ADA methodology, we substitute the conventional CNN-based feature extractor in ADA with the vanilla ViT to assess its efficacy. More specifically, we employ ViT as a feature extractor in DANN to learn domain-invariant features. To further explore the feasibility of ViT as a plug-and-play component in ADA, we also implement a variant of VT-ADA within the CDAN framework, thereby broadening the scope of our investigation into the adaptability of ViT in adversarial settings.

\section{Experiments}
\label{sec:Exp}

\begin{table}
\caption{Accuracy (\%) on the Office-31.}
\label{table1}
    \centering
    \resizebox{1.0\linewidth}{!}{
    \smallskip \begin{tabular}{cccccccc}
    \toprule[1pt]
    	Method & $\mathrm{A} \rightarrow \mathrm{W}$ & $\mathrm{D} 
	\rightarrow \mathrm{W}$ & $\mathrm{W} \rightarrow \mathrm{D}$ & $A 
	\rightarrow D$ & $\mathrm{D} \rightarrow \mathrm{A}$ & $\mathrm{W} 
	\rightarrow \mathrm{A}$ & Avg \\
	\hline 
	 DANN \cite{ganin2016domain} & $82.0 \pm 0.4$ & $96.9 \pm 0.2$ & $99.1 \pm 0.1$ & 
	$79.7 \pm 0.4$ & $68.2 \pm 0.4$ & $67.4 \pm 0.5$ & 82.2 \\
	 $\operatorname{ADDA}\cite{tzeng2017adversarial}$ & $86.2 \pm 0.5$ & $96.2 \pm 0.3$ & $98.4 
	\pm 0.3$ & $77.8 \pm 0.3$ & $69.5 \pm 0.4$ & $68.9 \pm 0.5$ & 82.9 \\
	 $\operatorname{CDAN}\cite{long2018conditional}$ & $93.1 \pm 0.2$ & $98.2 \pm 0.2$ & 
	$100.0 \pm 0.0$ & $89.8 \pm 0.3$ & $70.1 \pm 0.4$ & $68.0 \pm 0.4$ & 86.6 \\
	 $\mathrm{CDAN}+\mathrm{E}\cite{long2018conditional}$ & $94.1 \pm 0.1$ & $98.2 \pm 0.2$ & 
	$100.0 \pm 0.0$ & $92.9 \pm 0.2$ & $71.0 \pm 0.3$ & $69.3 \pm 0.4$ & 87.7 \\
    ETD \cite{li2020enhanced} & 92.1 & \textbf{100.0} & \textbf{100.0} & 88.0 & 71.0 & 67.8 & 86.2 \\
    CDTrans \cite{xu2021cdtrans} & \textbf{96.7} & 99.0 & \textbf{100.0} & \textbf{97.0} & 81.1 & \textbf{81.9} & \textbf{92.6} \\
	VT-ADA(DANN) & $95.8 \pm 0.3 $ & $98.3 \pm 0.3$ & $100.0 \pm \mathbf{0.0}$ & $91.3 \pm 0.2$ & $80.5 \pm 0.4$ & $80.2 \pm 0.4$ & 91.0 \\
	\textbf{VT-ADA(CDAN)} & $96.3\pm0.3$ & $99.1\pm0.1$ & $\mathbf{100.0}\pm\mathbf{0.0}$ & $93.1\pm0.4$ & $\mathbf{81.5} \pm \mathbf{0.2}$ & $78.9\pm0.2$ & 91.5\\
    \bottomrule[1pt]     
    \end{tabular}}
\end{table}


\begin{table}
\caption{Accuracy (\%) on the ImageCLEF.}
\label{table2}
    \centering
    \resizebox{1.0\linewidth}{!}{
    \smallskip \begin{tabular}{cccccccc}
    \toprule[1pt]
    Method & $\mathrm{I} \rightarrow \mathrm{P}$ & $\mathrm{P} 
	\rightarrow \mathrm{I}$ & $\mathrm{I} \rightarrow \mathrm{C}$ & 
	$\mathrm{C} \rightarrow \mathrm{I}$ & $\mathrm{C} \rightarrow \mathrm{P}$ 
	& $\mathrm{P} \rightarrow \mathrm{C}$ & Avg \\
	\hline 
	DANN\cite{ganin2016domain} & $75.0 \pm 0.6$ & $86.0 \pm 0.3$ & $96.2 \pm 0.4$ & $87.0 \pm 
	0.5$ & $74.3 \pm 0.5$ & $91.5 \pm 0.6$ & 85.0 \\
	CDAN\cite{long2018conditional} & $76.7 \pm 0.3$ & $90.6 \pm 0.3$ & $97.0 \pm 0.4$ & $90.5 \pm 
	0.4$ & $74.5 \pm 0.3$ & $93.5 \pm 0.4$ & 87.1 \\
	CDAN + E\cite{long2018conditional} & $77.7 \pm 0.3$ & $90.7 \pm 0.2$ & $97.0 \pm 0.3$ & $91.3 \pm 
	0.3$ & $74.2 \pm 0.2$ & $94.3 \pm 0.3$ & 87.7 \\
    ETD \cite{li2020enhanced} & 81.0 & 91.7 & \textbf{97.9} & 93.3 & \textbf{79.5} & 95.0 & 89.7 \\
	VT-ADA(DANN) & $79.6 \pm 0.3$ & $93.0 \pm 0.4$ & $96.4 \pm 0.4$ & $85.6 \pm 
	0.4$ & $70.3 \pm 0.3$ & $95.2 \pm 0.2$  & $86.7$  \\
	\textbf{VT-ADA(CDAN)} & $\mathbf{8 1 . 8} \pm \mathbf{0 . 3}$ & $\mathbf{9 4 . 1} \pm 
	\mathbf{0 . 2}$ & $97.3 \pm 0.1$ & $\mathbf{9 6 . 
	2} \pm \mathbf{0 . 3}$ & $7 8 . 9 \pm 0 . 3$ & 
	$\mathbf{9 6 . 0} \pm \mathbf{0 . 1}$ & $\mathbf{9 0 . 7}$ \\
    \bottomrule[1pt]     
    \end{tabular}}
\end{table}


\begin{table}
\caption{Accuracy (\%) on Office-Home.}
\label{table3}
    \centering
    \resizebox{1.0\linewidth}{!}{
    \smallskip \begin{tabular}{c c c c c c c c c c c c c c}
    \toprule[1pt]
    Method & Ar$\rightarrow$Cl & Ar$\rightarrow$Pr & Ar$\rightarrow$Rw & Cl$\rightarrow$Ar & Cl$\rightarrow$Pr & Cl$\rightarrow$Rw & Pr$\rightarrow$Ar & Pr$\rightarrow$Cl & Pr$\rightarrow$Rw & Rw$\rightarrow$Ar & Rw$\rightarrow$Cl & Rw$\rightarrow$Pr & Avg \\
        \hline
         DANN\cite{ganin2016domain} & 45.6 & 59.3 & 70.1 & 47.0 & 58.5 & 60.9 & 46.1 & 43.7 & 
	68.5 & 63.2 & 51.8 & 76.8 & 57.6 \\
	 CDAN\cite{long2018conditional} & 49.0 & 69.3 & 74.5 & 54.4 & 66.0 & 68.4 & 55.6 & 48.3 & 
	75.9 & 68.4 & 55.4 & 80.5 & 63.8 \\
     CDAN+E\cite{long2018conditional} & 50.7 & 70.6 & 76.0 & 57.6 & 70.0 & 70.0 & 57.4 & 50.9 & 
	77.3 & 70.9 & 56.7 & 81.6 & 65.8 \\
     ETD \cite{li2020enhanced} & 51.3 & 71.9 & 85.7 & 57.6 & 69.2 & 73.7 & 57.8 & 57.2 & 79.3 & 70.2 & 57.5 & 82.1 & 67.3 \\
     CDTrans \cite{xu2021cdtrans} & \textbf{60.6} & 79.5 & 82.4 & 75.6 & 81.0 & 82.3 & 72.5 & 56.7 & 84.4 & 77.0 & 59.1 & 85.5 & 74.7 \\
	 VT-ADA(DANN) & 52.5 & 81.1 & 86.4 & 71.8 & 79.7 & 81.3 & 73.3 & 55.3 & 
	87.5 & 79.3 & 58.3 & 87.8 & 74.5 \\
	 \textbf{VT-ADA(CDAN)} & 57.9 & \textbf{86.6} & \textbf{87.7} & \textbf{81.1} & \textbf{84.9} & \textbf{86.8} & \textbf{80.7} & \textbf{60.1} & \textbf{89.5} & \textbf{82.3} & \textbf{60.1} & \textbf{90.0} & \textbf{79.0} \\
    \bottomrule[1pt]     
    \end{tabular}}
\end{table}

\subsection{Datasets}
\label{sec:data}

\textbf{Office-31} is a benchmark domain adaptation dataset containing images belonging to 31 classes from three domains: Amazon (A) with 2,817 images, which contains images downloaded from amazon.com, Webcam (W) with 795 images and DSLR (D) with 498 images, which contain images taken by web camera and DSLR camera with different settings, respectively. We evaluate all methods on six domain adaptation tasks: \textbf{A} $\rightarrow$ \textbf{W}, \textbf{D} $\rightarrow$ \textbf{W}, \textbf{W} $\rightarrow$ \textbf{D}, \textbf{A} $\rightarrow$ \textbf{D}, \textbf{D} $\rightarrow$ \textbf{A}, and \textbf{W} $\rightarrow$ \textbf{A}.

\noindent\textbf{ImageCLEF} is a benchmark dataset for ImageCLEF 2014 domain adaptation challenges, which contains 12 categories shared by three domains: Caltech-256 (C), ImageNet ILSVRC 2012 (I), and Pascal VOC 2012 (P). Each domain contains 600 images and 50 images for each category. The three domains in this dataset are of the same size, which is a good complementation of the Office-31 dataset where different domains are of different sizes. We build six domain adaptation tasks: \textbf{I} $\rightarrow$ \textbf{P}, \textbf{P} $\rightarrow$ \textbf{I}, \textbf{I} $\rightarrow$ \textbf{C}, \textbf{C} $\rightarrow$ \textbf{I}, \textbf{C} $\rightarrow$ \textbf{P}, and \textbf{P} $\rightarrow$ \textbf{C}.

\noindent\textbf{Office-Home} is a more challenging dataset for UDA, consisting of around 15500 images from 65 classes in office and home settings, forming four extremely distinct domains: Artistic images (Ar), Clip Art (Cl), Product images (Pr), and Real-World images (Rw). We evaluate our methods on all twelve transfer tasks: \textbf{Ar} $\rightarrow$ \textbf{Cl}, \textbf{Ar} $\rightarrow$ \textbf{Pr}, \textbf{Ar} $\rightarrow$ \textbf{Rw}, \textbf{Cl} $\rightarrow$ \textbf{Ar}, \textbf{Cl} $\rightarrow$ \textbf{Pr}, \textbf{Cl} $\rightarrow$ \textbf{Rw}, \textbf{Pr} $\rightarrow$ \textbf{Ar}, \textbf{Pr} $\rightarrow$ \textbf{Cl}, \textbf{Pr} $\rightarrow$ \textbf{Rw}, \textbf{Rw} $\rightarrow$ \textbf{Ar}, \textbf{Rw} $\rightarrow$ \textbf{Cl}, and \textbf{Rw} $\rightarrow$ \textbf{Pr}.

\subsection{Implementation Details}
\label{sec: Imp}

In this study, all experiments were conducted using the PyTorch framework. We employed the ViT-Base with a $16 \times 16$ input patch size (ViT-B/16) \cite{dosovitskiy2020image}, which was pre-trained on ImageNet-21K, as our feature extractor to learn domain-invariant features. The ViT-B/16 configuration comprises 12 transformer layers. For training, we utilized mini-batch stochastic gradient descent with a momentum of 0.9, applying a learning rate annealing strategy inspired by \cite{ganin2015unsupervised}. The learning rate was dynamically adjusted according to the formula $\eta_p=\frac{\eta_0}{(1+\theta p)^\beta}$, where $p$ represents the progression of training epochs normalized between [0,1]. The initial learning rate $\eta_0$ was set to 0.01, with $\theta=10$ and $\beta=0.75$, parameters fine-tuned to ensure optimal convergence and minimal error in the source domain. Furthermore, the domain adaptation parameter $\lambda_d$ was gradually increased from 0 to 1, using the schedule $\frac{1-\exp(-\delta p)}{1+\exp(-\delta p)}$ where $\delta=10$, to effectively balance the influence of domain adaptation during training.

\subsection{Comparison Methods}

We enhance the domain adversarial neural network (DANN) \cite{ganin2016domain} and conditional adversarial domain adaptation (CDAN) \cite{long2018conditional} frameworks by integrating the ViT-B/16 as the feature extractor. Our approach is evaluated against several state-of-the-art (SOTA) methods, including DANN \cite{ganin2016domain}, CDAN, Conditional Adversarial Domain Adaptation with Entropy Conditioning (CDAN+E) \cite{long2018conditional}, Adversarial Discriminative Domain Adaptation (ADDA) \cite{tzeng2017adversarial}, Enhanced Transport Distance (ETD) \cite{li2020enhanced}, and Cross-Domain Transformer (CDTrans) \cite{xu2021cdtrans}, to underscore the comparative effectiveness of our proposed method.

\subsection{Results}

The results for Office-31 are detailed in Table \ref{table1}, with baseline results sourced from original publications where applicable. These findings demonstrate that our VT-ADA significantly enhances the accuracy of both DANN and CDAN, delivering competitive performance relative to leading UDA methodologies. Specifically, VT-ADA boosts DANN by $8.8\%$ and CDAN by $4.9\%$. Further, results for ImageCLEF and Office-Home are presented in Tables \ref{table2} and \ref{table3}, respectively. Notably, our VT-ADA(CDAN) variant outperforms other methodologies on these platforms. On ImageCLEF, VT-ADA enhances DANN by $1.7\%$ and CDAN by $3.6\%$, while on Office-Home, the increases are $16.9\%$ for DANN and $15.2\%$ for CDAN, indicating substantial gains on these more rigorous benchmarks. Moreover, VT-ADA(CDAN) dominates in 4 of 6 domains on ImageCLEF and 11 of 12 domains on Office-Home, surpassing the second-place CDTrans by $4.3\%$ on the latter, confirming the robustness of our method in challenging environments.

\begin{figure}[htb]\label{fig11}
\begin{minipage}[b]{.48\linewidth}
  \centering
  \centerline{\includegraphics[width=4.0cm]{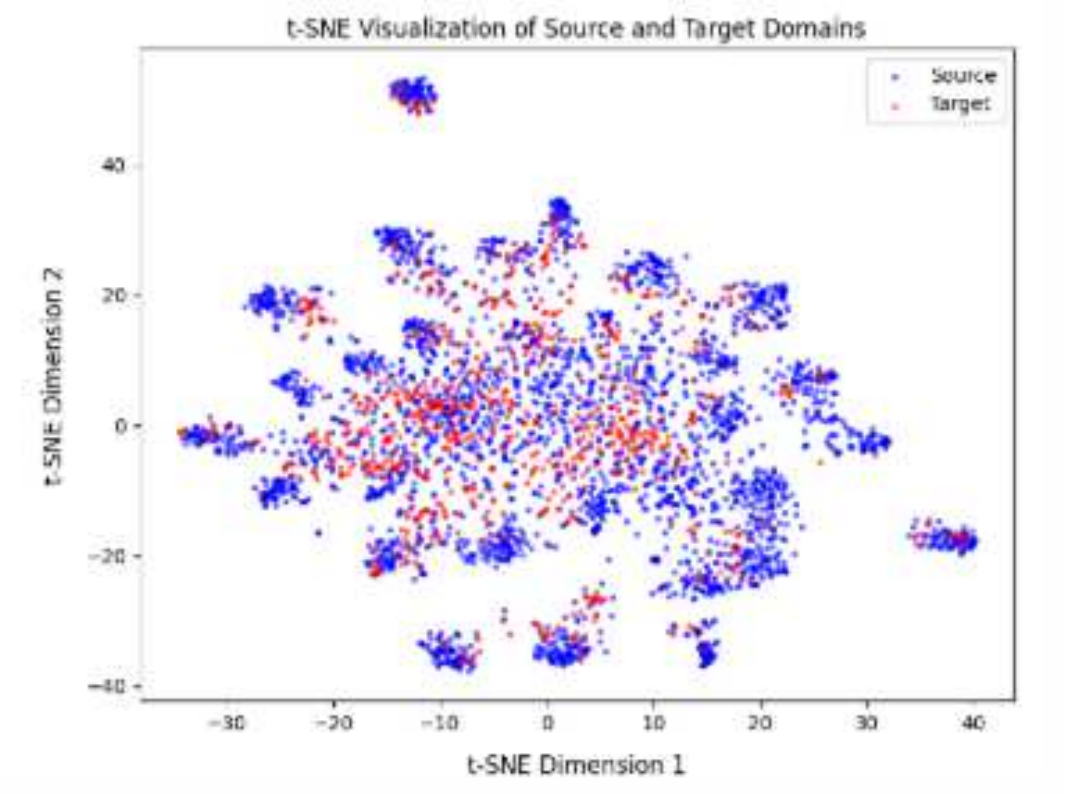}}
  \centerline{(b) DANN}\medskip
\end{minipage}
\hfill
\begin{minipage}[b]{0.48\linewidth}
  \centering
  \centerline{\includegraphics[width=4.0cm]{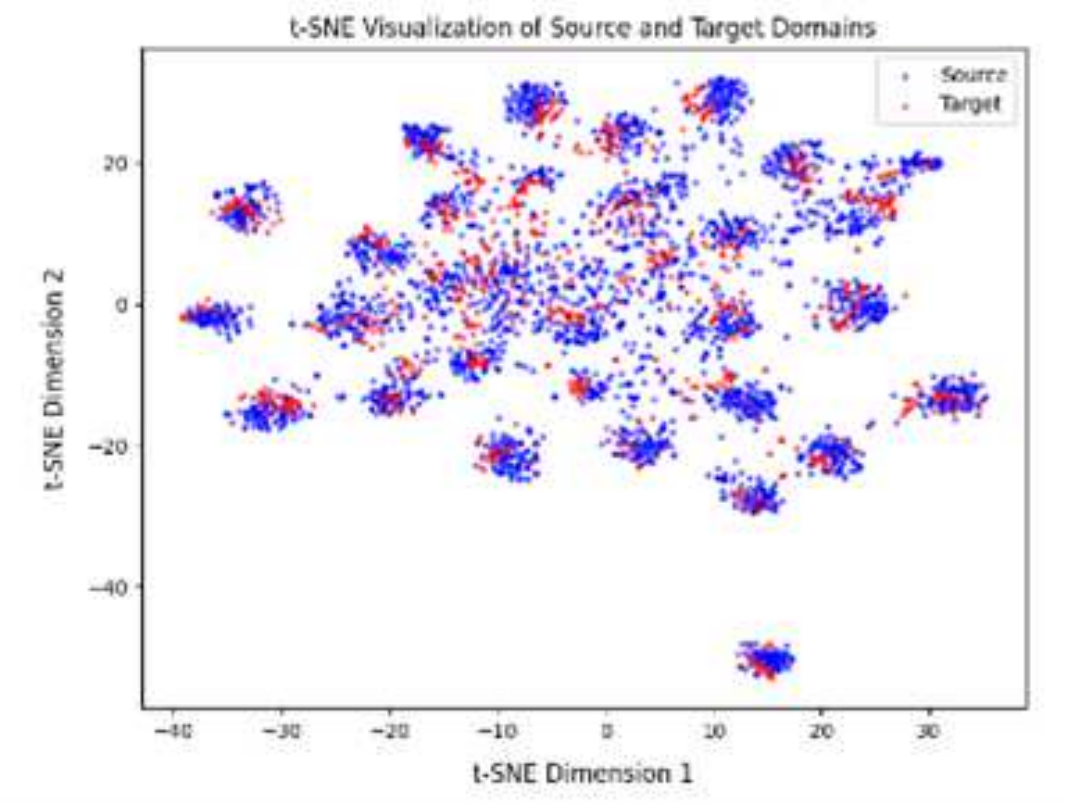}}
  \centerline{(c) CDAN}\medskip
\end{minipage}
\begin{minipage}[b]{.48\linewidth}
  \centering
  \centerline{\includegraphics[width=4.0cm]{CDAN.pdf}}
  \centerline{(b) CDAN+E}\medskip
\end{minipage}
\hfill
\begin{minipage}[b]{0.48\linewidth}
  \centering
  \centerline{\includegraphics[width=4.0cm]{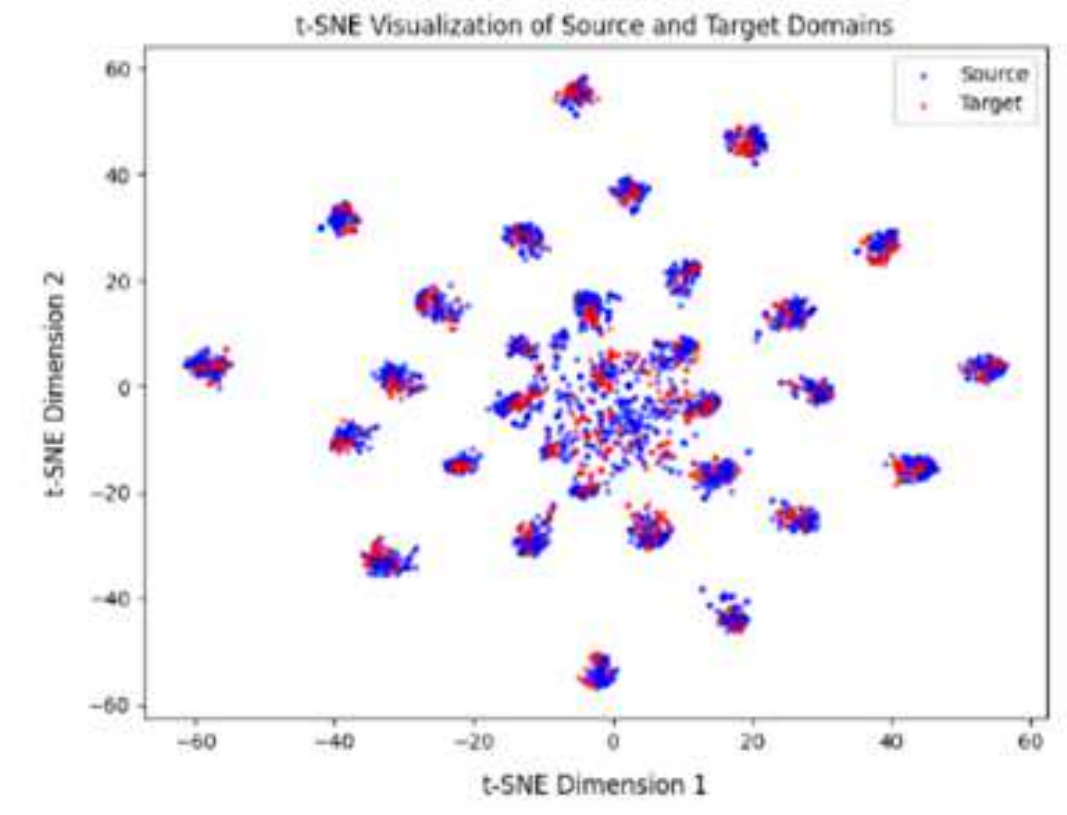}}
  \centerline{(c) VT-ADA(CDAN)}\medskip
\end{minipage}
	\caption{t-SNE\cite{van2008visualizing} visualization results for (a) DANN, (b) CDAN, (c) CDAN+E, 
	and (d) VT-ADA(CDAN).
		(Red points represent data points of domain A, while blue points represent data points of domain W)	 }	
\label{fig:res}
\end{figure}

\noindent\textbf{Visualization Analysis: } We employed the t-SNE technique to visualize the representations derived from DANN, CDAN, CDAN+E, and VT-ADA for task A → W of Office-31 dataset. The visualizations are displayed in Fig.1. Analysis of Fig.1 reveals that the feature alignments between source and target domains using DANN are suboptimal, with CDAN also demonstrating inadequate alignment. Conversely, CDAN+E shows improved alignment and enhanced classification performance. Most notably, the results from VT-ADA(CDAN) are markedly superior to those achieved with prior methods, underscoring the efficacy of employing ViT to learn more discriminative and transferable features within the context of UDA.

\begin{figure}[H]
    \centering
    \includegraphics[width=0.6\columnwidth]{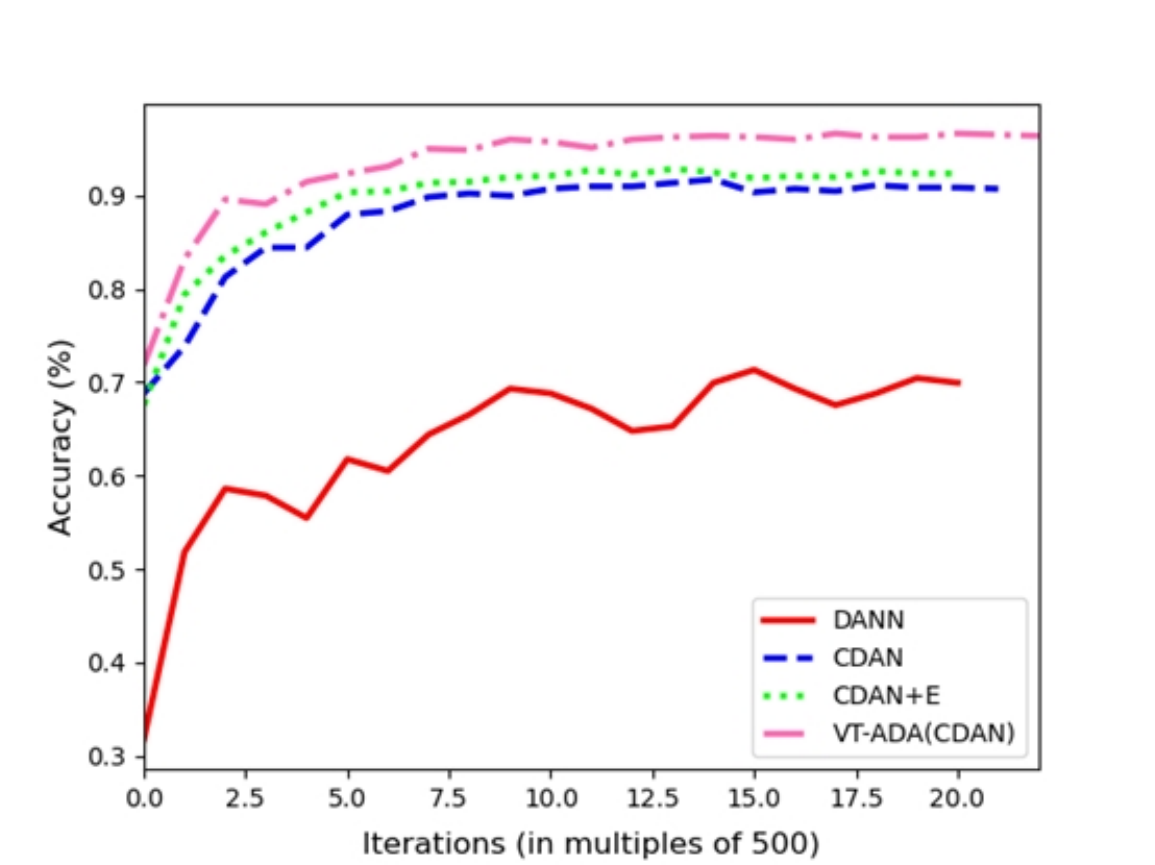}
    \caption{Convergence}
    \label{fig:enter-label}
\end{figure}

\noindent\textbf{Convergence Analysis:} We testify the convergence speed of DANN, CDAN, CDAN+E, and VT-ADA(CDAN), with the prediction accuracy on task A → W shown in Fig.2. It can be noteworthy that VT-ADA(CDAN) enjoys faster convergence speed among these methods.



\section{Conclusion}
\label{sec:con}

In this study, we propose a vision transformer-based adversarial domain adaptation (VT-ADA) method. This method utilizes ViT as a feature extractor to learn domain-invariant features. We empirically demonstrate that the ViT is an easy plug-and-play module that can be applied in existing ADA methods and bring advances. Additionally, our experiments on three UDA benchmarks and extensive analysis show that ViT has the ability to boost both the transferability and discriminability of the learned domain-invariant features and enhance the convergence speed of ADA methods.

\bibliographystyle{IEEEbib}
\bibliography{strings}

\begin{thebibliography}{10}

\bibitem{goodfellow2016deep}
Ian Goodfellow, Yoshua Bengio, and Aaron Courville,
\newblock {\em Deep learning},
\newblock MIT press, 2016.

\bibitem{han2022survey}
Kai Han, Yunhe Wang, Hanting Chen, Xinghao Chen, Jianyuan Guo, Zhenhua Liu,
  Yehui Tang, An~Xiao, Chunjing Xu, Yixing Xu, et~al.,
\newblock ``A survey on vision transformer,''
\newblock {\em IEEE transactions on pattern analysis and machine intelligence},
  vol. 45, no. 1, pp. 87--110, 2022.

\bibitem{pan2009survey}
Sinno~Jialin Pan and Qiang Yang,
\newblock ``A survey on transfer learning,''
\newblock {\em IEEE Transactions on knowledge and data engineering}, vol. 22,
  no. 10, pp. 1345--1359, 2009.

\bibitem{wang2018deep}
Mei Wang and Weihong Deng,
\newblock ``Deep visual domain adaptation: A survey,''
\newblock {\em Neurocomputing}, vol. 312, pp. 135--153, 2018.

\bibitem{gong2013connecting}
Boqing Gong, Kristen Grauman, and Fei Sha,
\newblock ``Connecting the dots with landmarks: Discriminatively learning
  domain-invariant features for unsupervised domain adaptation,''
\newblock in {\em International conference on machine learning}. PMLR, 2013,
  pp. 222--230.

\bibitem{yan2017mind}
Hongliang Yan, Yukang Ding, Peihua Li, Qilong Wang, Yong Xu, and Wangmeng Zuo,
\newblock ``Mind the class weight bias: Weighted maximum mean discrepancy for
  unsupervised domain adaptation,''
\newblock in {\em Proceedings of the CVPR}, 2017, pp. 2272--2281.

\bibitem{sun2016deep}
Baochen Sun and Kate Saenko,
\newblock ``Deep coral: Correlation alignment for deep domain adaptation,''
\newblock in {\em Computer Vision--ECCV 2016 Workshops: Amsterdam, The
  Netherlands, October 8-10 and 15-16, 2016, Proceedings, Part III 14}.
  Springer, 2016, pp. 443--450.

\bibitem{goodfellow2014generative}
Ian Goodfellow, Jean Pouget-Abadie, Mehdi Mirza, Bing Xu, David Warde-Farley,
  Sherjil Ozair, Aaron Courville, and Yoshua Bengio,
\newblock ``Generative adversarial nets,''
\newblock {\em Advances in neural information processing systems}, vol. 27,
  2014.

\bibitem{ganin2016domain}
Yaroslav Ganin, Evgeniya Ustinova, Hana Ajakan, Pascal Germain, Hugo
  Larochelle, Fran{\c{c}}ois Laviolette, Mario Marchand, and Victor Lempitsky,
\newblock ``Domain-adversarial training of neural networks,''
\newblock {\em The journal of machine learning research}, vol. 17, no. 1, pp.
  2096--2030, 2016.

\bibitem{long2018conditional}
Mingsheng Long, Zhangjie Cao, Jianmin Wang, and Michael~I Jordan,
\newblock ``Conditional adversarial domain adaptation,''
\newblock {\em Advances in neural information processing systems}, vol. 31,
  2018.

\bibitem{li2021survey}
Zewen Li, Fan Liu, Wenjie Yang, Shouheng Peng, and Jun Zhou,
\newblock ``A survey of convolutional neural networks: analysis, applications,
  and prospects,''
\newblock {\em IEEE transactions on neural networks and learning systems},
  2021.

\bibitem{dosovitskiy2020image}
Alexey Dosovitskiy, Lucas Beyer, Alexander Kolesnikov, Dirk Weissenborn,
  Xiaohua Zhai, Thomas Unterthiner, Mostafa Dehghani, Matthias Minderer, Georg
  Heigold, Sylvain Gelly, et~al.,
\newblock ``An image is worth 16x16 words: Transformers for image recognition
  at scale,''
\newblock {\em arXiv preprint arXiv:2010.11929}, 2020.

\bibitem{liu2022swin}
Ze~Liu, Han Hu, Yutong Lin, Zhuliang Yao, Zhenda Xie, Yixuan Wei, Jia Ning, Yue
  Cao, Zheng Zhang, Li~Dong, et~al.,
\newblock ``Swin transformer v2: Scaling up capacity and resolution,''
\newblock in {\em Proceedings of the CVPR}, 2022, pp. 12009--12019.

\bibitem{li2022exploring}
Yanghao Li, Hanzi Mao, Ross Girshick, and Kaiming He,
\newblock ``Exploring plain vision transformer backbones for object
  detection,''
\newblock in {\em European Conference on Computer Vision}. Springer, 2022, pp.
  280--296.

\bibitem{strudel2021segmenter}
Robin Strudel, Ricardo Garcia, Ivan Laptev, and Cordelia Schmid,
\newblock ``Segmenter: Transformer for semantic segmentation,''
\newblock in {\em Proceedings of the ICCV}, 2021, pp. 7262--7272.

\bibitem{zhou2024survey}
Yue Zhou, Chenlu Guo, Xu~Wang, Yi~Chang, and Yuan Wu,
\newblock ``A survey on data augmentation in large model era,''
\newblock {\em arXiv preprint arXiv:2401.15422}, 2024.

\bibitem{wu2020dual}
Yuan Wu, Diana Inkpen, and Ahmed El-Roby,
\newblock ``Dual mixup regularized learning for adversarial domain
  adaptation,''
\newblock in {\em Computer Vision--ECCV 2020: 16th European Conference,
  Glasgow, UK, August 23--28, 2020, Proceedings, Part XXIX 16}. Springer, 2020,
  pp. 540--555.

\bibitem{tzeng2017adversarial}
Eric Tzeng, Judy Hoffman, Kate Saenko, and Trevor Darrell,
\newblock ``Adversarial discriminative domain adaptation,''
\newblock in {\em Proceedings of the CVPR}, 2017, pp. 7167--7176.

\bibitem{xu2021cdtrans}
Tongkun Xu, Weihua Chen, WANG Pichao, Fan Wang, Hao Li, and Rong Jin,
\newblock ``Cdtrans: Cross-domain transformer for unsupervised domain
  adaptation,''
\newblock in {\em International Conference on Learning Representations}, 2021.

\bibitem{vaswani2017attention}
Ashish Vaswani, Noam Shazeer, Niki Parmar, Jakob Uszkoreit, Llion Jones,
  Aidan~N Gomez, {\L}ukasz Kaiser, and Illia Polosukhin,
\newblock ``Attention is all you need,''
\newblock {\em Advances in neural information processing systems}, vol. 30,
  2017.

\bibitem{chang2023survey}
Yupeng Chang, Xu~Wang, Jindong Wang, Yuan Wu, Linyi Yang, Kaijie Zhu, Hao Chen,
  Xiaoyuan Yi, Cunxiang Wang, Yidong Wang, et~al.,
\newblock ``A survey on evaluation of large language models,''
\newblock {\em ACM Transactions on Intelligent Systems and Technology}, 2023.

\bibitem{li2020enhanced}
Mengxue Li, Yi-Ming Zhai, You-Wei Luo, Peng-Fei Ge, and Chuan-Xian Ren,
\newblock ``Enhanced transport distance for unsupervised domain adaptation,''
\newblock in {\em Proceedings of the CVPR}, 2020, pp. 13936--13944.

\bibitem{ganin2015unsupervised}
Yaroslav Ganin and Victor Lempitsky,
\newblock ``Unsupervised domain adaptation by backpropagation,''
\newblock in {\em International conference on machine learning}. PMLR, 2015,
  pp. 1180--1189.

\bibitem{van2008visualizing}
Laurens Van~der Maaten and Geoffrey Hinton,
\newblock ``Visualizing data using t-sne.,''
\newblock {\em Journal of machine learning research}, vol. 9, no. 11, 2008.

\end{thebibliography}

\end{document}